\documentclass[10pt,twocolumn,letterpaper]{article}

\usepackage{iccv}
\usepackage{times}
\usepackage{epsfig}
\usepackage{graphicx}
\usepackage{amsmath}
\usepackage{amssymb}
\usepackage[american]{babel}
\graphicspath{{figure/}}

\usepackage[breaklinks=true,bookmarks=false]{hyperref}

\iccvfinalcopy 

\def\iccvPaperID{3731} 

\newcommand{\squishlist}{
	\begin{list}{$\bullet$}
		{ \setlength{\itemsep}{0pt}
			\setlength{\parsep}{1pt}
			\setlength{\topsep}{1pt}
			\setlength{\partopsep}{0pt}
			\setlength{\leftmargin}{1.5em}
			\setlength{\labelwidth}{1em}
			\setlength{\labelsep}{0.5em} } }
\newcommand{\squishend}{
\end{list}  }
\begin{document}

\title{Towards Real-Time Action Recognition on Mobile Devices Using Deep Models}
\author{Chen-Lin Zhang, Xin-Xin Liu, and Jianxin Wu\\
	National Key Laboratory for Novel Software Technology\\ Nanjing University, Nanjing, China\\
	{\tt\small \{zclnjucs, xinxliu1996, wujx2001\}@gmail.com}
}
\maketitle

\begin{abstract}
   Action recognition is a vital task in computer vision, and many methods are developed to push it to the limit. However, current action recognition models have huge computational costs, which cannot be deployed to real-world tasks on mobile devices. In this paper, we first illustrate the setting of real-time action recognition, which is different from current action recognition inference settings. Under the new inference setting, we investigate state-of-the-art action recognition models on the Kinetics dataset empirically. Our results show that designing efficient real-time action recognition models is different from designing efficient ImageNet models, especially in weight initialization. We show that pretrained weights on ImageNet improve the accuracy under the real-time action recognition setting. Finally, we use the hand gesture recognition task as a case study to evaluate our compact real-time action recognition models in real-world applications on mobile phones. Results show that our action recognition models, being 6x faster and with similar accuracy as state-of-the-art, can roughly meet the real-time requirements on mobile devices. To our best knowledge, this is the first paper that deploys current deep learning action recognition models on mobile devices.
\end{abstract}

\section{Introduction}
Video-based action recognition, which is a vital task in computer vision, has drawn enormous attention from the community~\cite{idt2013iccv,zhu2011wearable,resnetcvpr2016,li2018resound}. Since the remarkable success of deep learning, researchers have proposed various deep models for action recognition. These models need enormous computational resources and are mostly applied on GPU equipped servers.

However, due to the rapid growth of mobile applications, there are increasing demands for conducting action recognition tasks on mobile devices. Some researches have been done on performing single image recognition on portable devices directly~\cite{mobilenetv2cvpr2018,shufflenetcvpr2018}, and some other researches try to collect sensor's data to conduct action recognition, which does not use the camera sensor. However, temporal dependency plays a vital role in these tasks, and it is insufficient to only conduct single image and sensor data based action recognition. Due to the large video file size, it is not applicable to transfer them to a cloud server with GPUs for processing. Hence, it is essential to perform action recognition tasks on mobile devices directly. 

In action recognition, many methods try to explain and capture the temporal dependencies between frames and push the limit of current state-of-the-art models. 3D convolutional models are proposed to capture the dense temporal dependencies in the time domain~\cite{c3d2015cvpr,res3d2017arxiv,artnetcvpr2018,mfnet2018eccv,resnext3d2018cvpr}. Besides 3D models, some other models are proposed to capture the long-term temporal dependencies in videos~\cite{tsn2016eccv,trn2018eccv}. However, all these deep learning methods need enormous computational resources~(hundreds of BFLOPs and millions of parameters).

Many architectures have been developed to conduct image recognition tasks on mobile platforms~\cite{mobilenetv2cvpr2018,mnasnet2018arxiv}. However, little attention has been paid on developing efficient computational action recognition models, especially in a resource constrained environment like mobile devices. Current state-of-the-art action recognition models are too heavy to run on mobile devices for real-world applications. Hence, currently mobile devices still use non-deep learning methods to conduct action recognition tasks, whose accuracy are lower than the state-of-the-art deep learning methods.

Besides the architecture part, the inference setting of performing action recognition on mobile devices should be different from current inference setting of many action recognition models. State-of-the-art action recognition models will perform heavy data augmentations from input videos when testing, e.g., sample 10 or more clips from the original videos and 10 crop image augmentation, then average the predictions among all these clips. However, heavy data augmentation will consume huge preprocessing time and cannot meet the low latency requirement of real-time action recognition. Hence, heavy data augmentation is not preferred on mobile devices. 

All these factors encourage us to explore the inference setting of action recognition on mobile devices. We re-evaluate current state-of-the-art models under this new inference setting, and achieve state-of-the-art performance under the real-time constraint. Based on these new results, we have some suggestions for future model design. 

Our contributions are listed as follows:
\squishlist
	\item We explore the setting of real-time action recognition tasks on devices with limited resources like mobile phones, and perform an empirical study of state-of-the-art action recognition models on the Kinetics dataset under the real-time action recognition setting. 
	\item With the help of some extra modules, we achieve performance comparable to state-of-the-art models with 6x fewer FLOPs and running time on the Kinetics dataset.
	\item We conduct a case study on the Jester gesture recognition dataset in mobile environments, which can lead to efficient action recognition models in real-world tasks. To our best knowledge, we are the first to introduce deep learning action recognition models onto mobile devices for real-time processing.    
\squishend
For efficient action recognition model design, our results show that designing efficient real-time action recognition models is different from designing image recognition models. In detail, we have the following findings:
\squishlist
	
	\item In the real-time action recognition setting, 2D based models with temporal segments~(TSN) are superior to 3D conv and other models considering both FLOPs and accuracy.
	\item In the real-time action recognition setting, models with joint training on ImageNet and Kinetics perform better than training only on Kinetics. Moreover, the accuracy on Kinetics is directly correlated with accuracy on ImageNet, which is different from previous conclusions.
	\item For compact models, overfitting is a key factor that hinders performance, which is also different from ImageNet conclusions.
	\item Branch architectures like Inception are suitable for the real-time action recognition tasks, but it will also cost a lot of mobile CPU latency, even with the same FLOPs.
\squishend

\section{Related Works}

In this section, we will give a brief introduction to current deep learning models and action recognition models.
\subsection{Efficient Deep Learning}

Efficient deep learning models have drawn a lot of attention in the computer vision society. Early network designs such as VGGNet{~\cite{vggiclr2015}} and ResNet~\cite{resnetcvpr2016} are proven to be redundant, and pruning methods are proposed to prune these models~\cite{thinet17,nispcvpr2018}. Other researchers focus on directly designing new efficient modules and structures, e.g., MobileNet series~\cite{mobilenetv2cvpr2018} and ShuffleNet~\cite{shufflenetcvpr2018}. 

Besides these hand-crafted neural architectures, NAS~(Neural Architecture Search), which focuses on developing methods to generate efficient neural network architectures automatically, become popular these days. The pioneering work~\cite{nas2017iclr} uses reinforcement learning algorithms to train a Recurrent Neural Network~(RNN) controller that generates coded architectures. Some works try to accelerate the whole search process with regularization and pruning. NASNet~\cite{nasnetcvpr2018} searches architectures on CIFAR-10 and transfer the searched architectures to the large-scale ImageNet. PNASNet~\cite{pnasneteccv2018} factorizes the search space and accelerates the search process, achieving comparable results with NASNet. MnasNet~\cite{mnasnet2018arxiv} takes real-world device latency into the searching constraint.

All these efficient models are designed for image recognition, and some works transfer the image recognition model to other tasks like object detection and semantic segmentation~\cite{mobilenetv2cvpr2018}. However, little effort has been paid on developing efficient models in action recognition tasks, which encourages us to explore this new problem setting.

\subsection{Action Recognition}
\begin{figure*}
	\centering
	\includegraphics[width=0.9\linewidth]{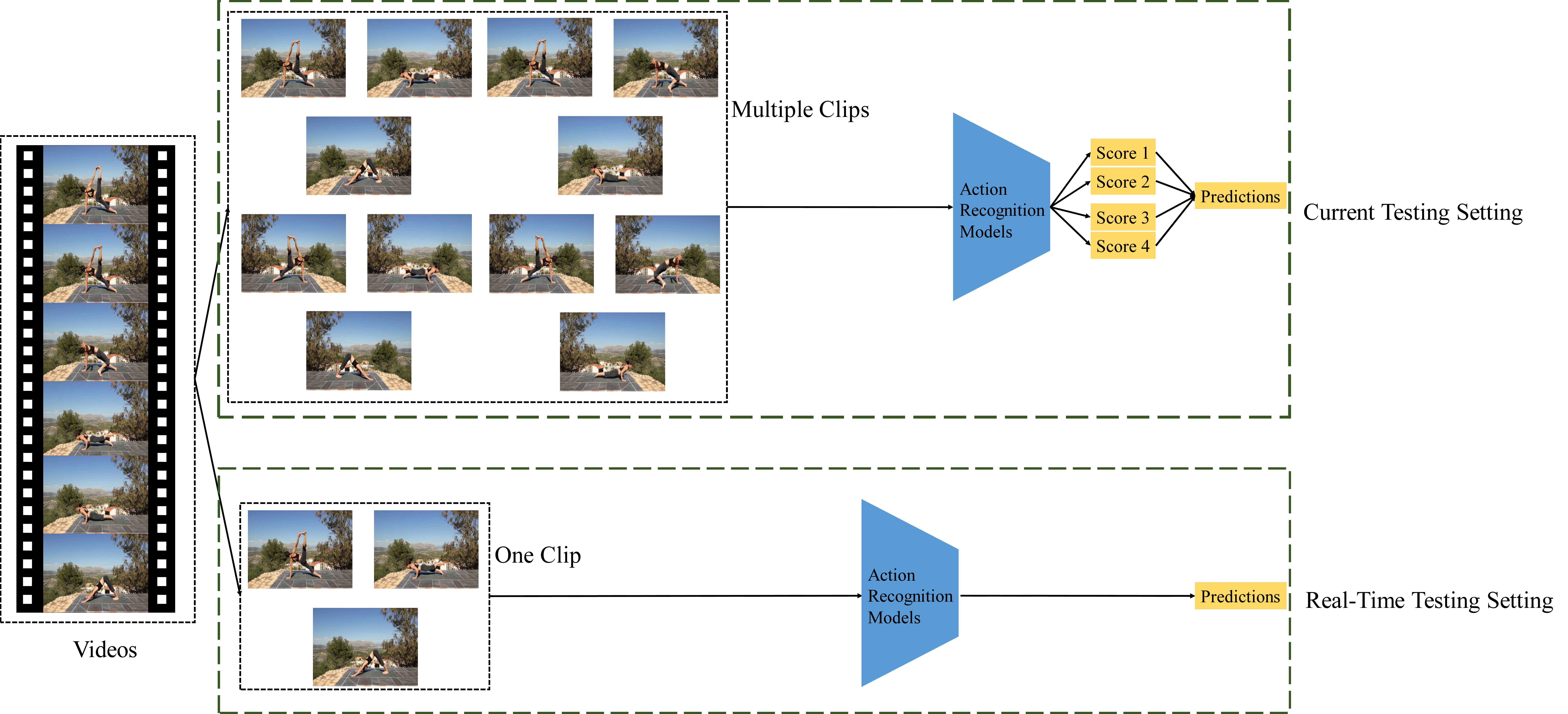}
	\caption{Overall inference framework of two action recognition settings. The key differences between these two settings are the data augmentation and size of the models. }\label{figure: overall}
	
\end{figure*}
Compared to the image recognition task, action recognition is much harder because there exist time dependencies in videos. Before the era of deep learning, various hand-crafted features are proposed~\cite{idt2013iccv}. Motion vectors~(optical flows) are also proved to be useful~\cite{op1981ijcai}. After the success of deep learning, researchers start to use deep learning in action recognition tasks.

Two-Stream ConvNet~\cite{twostream2014nips} is a popular method in the field of action recognition, which uses both RGB and optical flow modality. Some improvements are proposed then to achieve better results~\cite{twostreami2016cvpr}. However, methods in this two stream fashion pay no attention to the time axes and need extra modalities.

Another popular trend is to expand spatial~(2D) CNN into spatial-temporal~(3D) CNN. However, 3D CNNs have more parameters than 2D CNNs, and they consume huge resources in the training and testing process. Meanwhile, since most datasets in action recognition are small~(about 10k videos), 3D CNNs cause serious overfitting problems. In 2017, the Kinetics dataset~\cite{kinetics2017cvpr}, which has about 300k videos, is proposed to solve the dataset problem in action recognition. Since the publishing of Kinetics, various 3D models are proposed to improve the performance~\cite{stcnn2018eccv,iasteccv2018}. ~\cite{kinetics2017cvpr} first proposes I3D network, which inflates 2D convolution kernels into 3D convolution kernels. ~\cite{s3d2018eccv} shows that 3D convolution near the classifier is helpful for action recognition. ~\cite{mfnet2018eccv} uses group convolution to enlarge the capacity of 3D CNN models. However, these models can only take short clips due to memory limits and lack attention in long-term dependencies. 

Again, little attention has been paid on applying current action recognition models on real-world tasks. Some efforts have been paid on online video understanding~\cite{eco2018eccv,tsmarxiv2018}. However, both are applied on GPU devices, without trying to apply it on mobile devices.

\section{A Realistic Setting}
The key difference between current inference setting and real-time action recognition setting is the running time. The goal of current state-of-the-art inference setting is achieving high accuracy. However, real-time action recognition needs to consider the inference time constraint.

To illustrate the real-time action recognition better, we list the environment constraint below~(cf. Fig.~\ref{figure: overall}):

\squishlist
	\item Heavy data augmentation is not possible: Data augmentation is not preferred especially in mobile environments due to the long processing time.
	\item Small model size and FLOPs: Mobile devices have much fewer computation and storage resources than desktop CPUs and GPUs. Since we need to input multiple frames into the models, we need models with fewer parameters and FLOPs.
	\item Extra modality is not preferred: We need low preprocessing time. Hence, we can not add extra modalities like optical flow and RGB difference. 
\squishend

Based on these constraints, we propose the following setting for real-time action recognition:

``Given an input video or a stream of input frames, we directly sample frames and do not perform any data augmentations. Then, these sampled frames are fed into the model, and the predictions are obtained through its output.'' 

This setting is different from most inference protocols of current action recognition models. Current inference protocols mostly use heavy data augmentations like 10 crop and 10 sampled clips from original videos and they often use these results to compare with other methods. However, these results are unsuitable in this real-time problem setting. Some 3D models provide results such as clip@1, and these results use the same inference protocol as our real-time action recognition protocol. We will directly cite these results in our paper. However, for other methods, they do not report results in this new setting. We need to re-evaluate the performance of state-of-the-art models. 

\section{Experimental Configuration}

In this section, we will give detailed experimental setups.
\subsection{Base Model Preparation}
\begin{figure*}
	\includegraphics[width=1\linewidth]{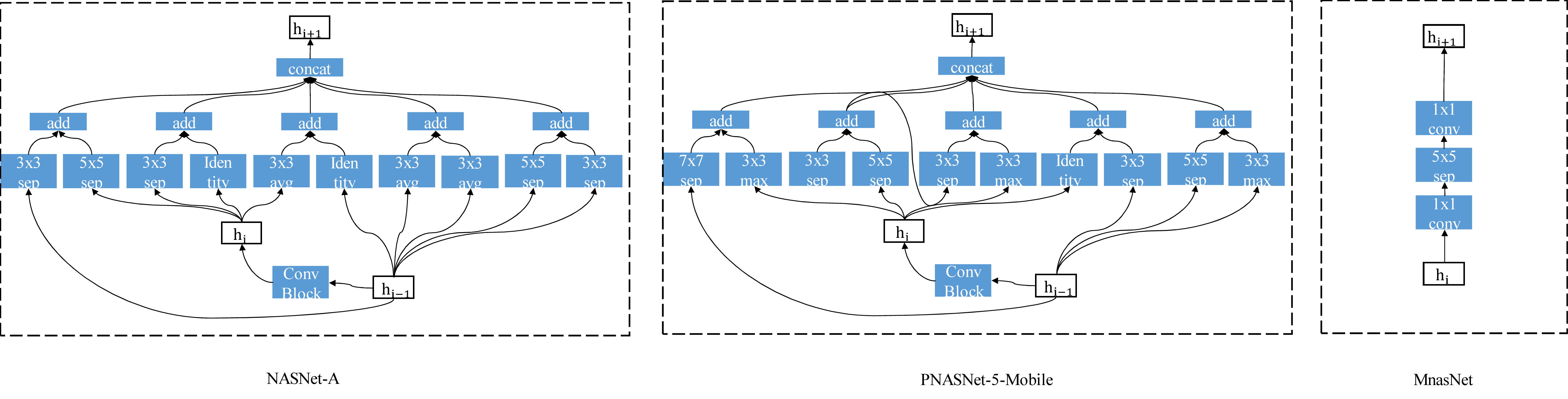}
	\caption{Architectures of exactly one building cell of the three NAS searched models. Please note that in this figure, \texttt{avg} and \texttt{max} represent average and maximum pooling, \texttt{sep} represents   depthwise convolution operation. }\label{figure: nas}	
\end{figure*}
For empirically studying the effect of different modules and architectures, we prepare these models:
\squishlist
	\item Compressed VGG models~(ThiNet-Tiny): Various techniques are developed for compressing CNNs. Among these models, ThiNet achieves state-of-the-art~(better than AlexNet) performance with only 1.32M parameters~\cite{thinet17}.
	\item MobileNet series~(MobileNet V1 and MobileNet V2): The MobileNet series of models are created to conduct deep learning tasks on mobile devices. MobileNetV1 uses depthwise separable convolution to reduce computation costs~\cite{mobilenetv2cvpr2018}. MobileNetV2 adds inverted residual blocks and residual connections to the MobileNet and improves the performance~\cite{mobilenetv2cvpr2018}. These two models achieve good results on ImageNet. For further comparison, we add an extra widen MobileNetV2 model called MobileNetV2-1.4, which has about 2x FLOPs than MobileNetV2 and 2.7\% performance gain on ImageNet.
	\item NASNet-A-mobile: After the pioneering work on NAS, some improvements are made to transfer the learned network structure from CIFAR-10 to large-scale datasets like ImageNet~\cite{nasnetcvpr2018}. NASNet-A-mobile, which achieves 74.0\% accuracy on ImageNet with only 5.3M parameters and 564M FLOPs, is also a good backbone network for action recognition.
	\item PNASNet-5-mobile: Compared to NASNet~\cite{nasnetcvpr2018}, PNASNet uses progressive search strategies to reduce search costs for NASNet with comparable costs~\cite{pnasneteccv2018}. The mobile series of PNASNet, PNASNet-5-mobile, achieves comparable results with NASNet.
	\item MnasNet: After NASNet, MnasNet~\cite{mnasnet2018arxiv} uses reinforcement learning to take real CPU latency on mobile phones into consideration and achieves comparable results with NASNet-A-mobile with less real latency on mobile devices.
\squishend

For the first three manually designed networks, their building blocks are rather simple. They are mostly composed of simple convolution layers~(including traditional, depthwise and separable convolution). For NAS searched models, basic blocks of these networks are shown in Fig.~\ref{figure: nas}. From these figures, we can see that NAS models are more complex than manually designed models, except for MnasNet. 

\subsection{2D or 3D CNN models?}
Long-term dependency is another critical factor for action recognition. Some previous methods use LSTM to handle long-term dependencies~\cite{lstmcvpr2015}. However, these methods also have overfitting problems.

Temporal segment network~(TSN)~\cite{tsn2016eccv} shows that random sampling between large intervals can also capture the long-term dependencies in a video. ~\cite{trn2018eccv} proposes a relational reasoning module to capture dependencies better. In this section, we will give a brief recapitulation to TSN. 

The TSN method is composed of three parts: A random sparse sampling part, a spatial ConvNet part, and an aggregation part. Given a video $\mathbf{V}$, TSN will divide $\mathbf{V}$ into $k$ segments: $\{Seg_1,\ldots,Seg_k\}$. Then precisely one frame will be sampled from each segment. These sampled images will be fed into a spatial CNN to get independent predict scores for each frame. Finally, these scores are fused to get the final prediction score for the whole video. In TSN, the author chooses random sampling techniques for training TSN-based models and medium sampling for inference. For aggregating function, the author chooses the \emph{average} function, i.e., all predictions are averaged together to get the final prediction. 

For base method choice and comparison, there is a fundamental question: Should we choose 3D conv-based models, or do we need to choose methods based on 2D models?

We conducted experiments on 2D and 3D CNN models. For all models with TSN, we add a dropout layer with 0.8 dropout ratio, following the suggestions in~\cite{tsn2016eccv}. According to Table~\ref{table:2d3d results} and Fig.~\ref{figure: Flops}, we can easily see that:
\begin{figure*}
	\centering
	\includegraphics[width=0.9\linewidth]{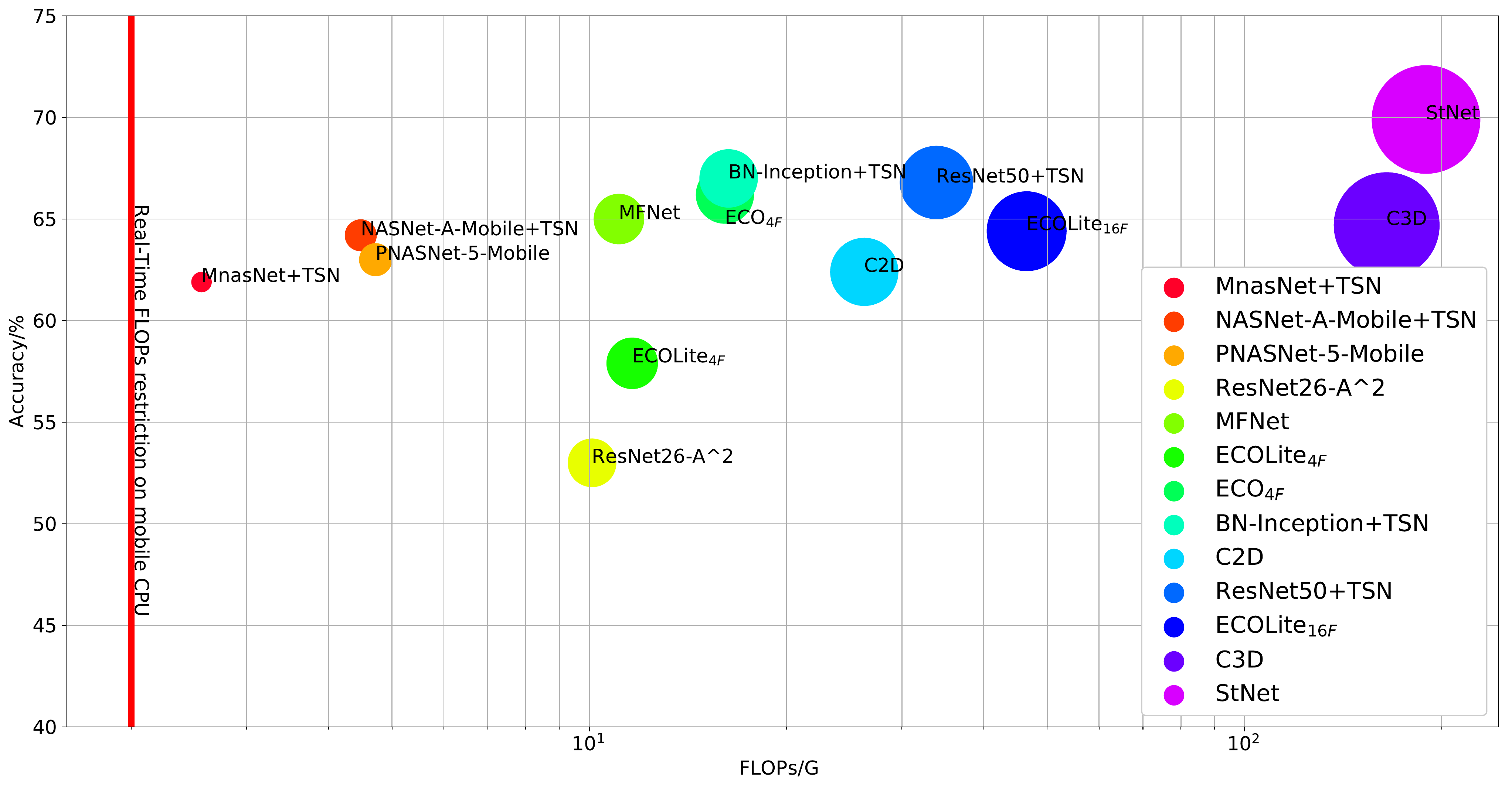}
	\caption{Accuracy-FLOPs trade-off on Kinetics validation set for various versions of TSN and other state-of-the-art approaches. Please note that all results are tested under the inference setting of real-time action recognition. 2D models with TSN are much closer to the top left corner than any other approach. For FLOPs calculation, we feed the sampled clip into models and calculate the overall FLOPs. The line in this figure represent to the FLOPs upper bound of current mobile CPUs.}\label{figure: Flops}
	
\end{figure*}
In the setting of real-time action recognition, 3D models are inferior to 2D based models. With 11.1B FLOPs, the state-of-the-art MFNet~\cite{mfnet2018eccv} only achieves 65.0\% accuracy on Kinetics. However, NASNet-A-Mobile with TSN achieves 64.2\% accuracy with only 4.48B FLOPs, which is about two times less than MFNet. For other 3D models, the FLOPs is mostly over 100G and is not suitable for running on mobile devices. Furthermore, 2D models with TSN performs much better than single 2D models. From Fig.~\ref{figure: Flops}, we can see that 2D models with TSN are in the up left corner. According to the observations, we decide to use 2D models with TSN in our subsequent experiments. 
\begin{table}
	
	\begin{small}
		\begin{center}
			\caption{Performance of state-of-the-art models~(number of parameters, FLOPs and accuracy) of current 2D and 3D models re-evaluated under the real-time action recognition setting on the Kinetics dataset. The second column is the number of input frames and the third to fifth column represent the number of parameters, FLOPs and accuracy on the Kinetics validation dataset. Please note that all models only take 1 clip from the RGB modality of origin video and all results with our TSN use models initialized with the weights pretrained from ImageNet. Some results are drawn from previous papers~\cite{tsmarxiv2018,a2netnips2018,stnetaaai2019,eco2018eccv}.}
			\label{table:2d3d results}
			\setlength{\tabcolsep}{2pt}
			\begin{tabular}{|l|r|r|r|r|}
				\hline
				Model & Frames & Params & FLOPs & Acc\\
				\hline
				C2D~\cite{artnetcvpr2018} & 32 & 24.3M & 26.3B & 62.4\% \\
				BN-Inception+TSN~\cite{tsn2016eccv} & 8 & 11.3M  &16.3B& 67.0\%  \\
				ResNet-50+TSN~\cite{tsn2016eccv} & 8 & 25.6M & 33.9B & 66.8\% \\
				\hline
				MnasNet+TSN~\cite{mnasnet2018arxiv} & 8 & 5.1M & 2.6B & 61.9\% \\
				NASNet-A-Mobile+TSN~\cite{nasnetcvpr2018} & 8 & 4.2M  &4.5B& 64.2\% \\
				PNASNet-5-Mobile+TSN~\cite{pnasneteccv2018} &8 & 5.1M  & 4.7B & 63.0\% \\
				\hline 
				C3D~\cite{artnetcvpr2018} & 32 & 35.0M & 164.8B & 64.7\%\\
				MFNet~\cite{mfnet2018eccv} & 16 & 8.0M & 11.1B & 65.0\% \\
				StNet~\cite{stnetaaai2019} & 75 & 33.2M & 189.3B & 69.9\% \\
				ResNet26+A$^2$~\cite{a2netnips2018} & 16 & 7.7M & 10.1B & 53.0\% \\
				ECOLite$_{4F}$~\cite{eco2018eccv} & 4 & 37.5M & 11.6B & 57.9\% \\
				ECOLite$_{16F}$~\cite{eco2018eccv} & 16 & 37.5M & 11.6B & 64.4\% \\
				ECO$_{4F}$~\cite{eco2018eccv} & 4 & 47.5M & 16.1B & 66.2\% \\ 
 				\hline
			\end{tabular}
		\end{center}
	\end{small}
	
\end{table}


\subsection{Experimental Setup}

We use PyTorch for our deep learning framework and the PyTorch implementation of TSN code.\footnote{https://github.com/yjxiong/tsn-pytorch} We use a GPU server equipped with eight M40 GPUs, two Intel E5-2650 v4 CPUs and 256GB memory support for training and testing our models. All our models are trained using single precision~(FP32). For all these models, we use MMdnn to convert them into PyTorch pretrained models and report their performance on ImageNet 2012 validation dataset. For inferring video on mobile devices, we use a HUAWEI Mate 10 phone with Kirin 970 CPU and 6GB RAM. Note that we do not use the neural network chip in Kirin to accelerate computing. For mobile platforms, we use the TFLite and related projects. 

For better evaluating our models, we choose the Kinetics dataset to evaluate our models. Kinetics is a large-scale high-quality dataset of YouTube videos, which include a diverse range of human-focused actions~\cite{kinetics2017cvpr}. Videos in Kinetics are clipped from Youtube videos, while some other datasets~(Chardes, Something-Something) are collected from the Amazon Mechanical Turk~(AMT). The Kinetics dataset is highly related to daily actions, which is similar to the tasks in mobile action recognition scenes. 

We evaluate our models on the trimmed version. In this paper, we use Kinetics-400. Kinetics-400 consists of approximately 300,000 video clips and covers 400 human action classes with at least 400 video clips for each action class. For validation, each class has 50 video clips. We remove some unavailable videos in Kinetics-400. There are about 240k videos in the training set and we report the accuracy on the validation dataset. 

Hyperparameters are set the same for all these experiments. The learning rate is $0.01$ for first 45 epochs, and decay 0.1 every 15 epochs. The weight decay is $0.0005$ and the batch size is 128. We clip all the gradients when the gradients are larger than 20. Unless specified, all model do not have the last dropout layer, i.e., the dropout ratio is set to 0 for all models.

For testing, we first sample 8 RGB frames from each action video. Different from standard TSN protocol, we do not perform 10 crop data augmentation among the data input due to the constraint we have mentioned before. Finally, 8 images are fed into the TSN models to get individual scores for each input. Then 8 scores are averaged to get the final score. 

\section{Results and Analysis}
In this section, we will provide detailed experimental results and findings.
\subsection{General results}
First, we conduct experiments over all base models. Please note that all our base models in this section use pretrained weights on ImageNet.
\begin{figure*}
	\centering
	\includegraphics[width=1\linewidth]{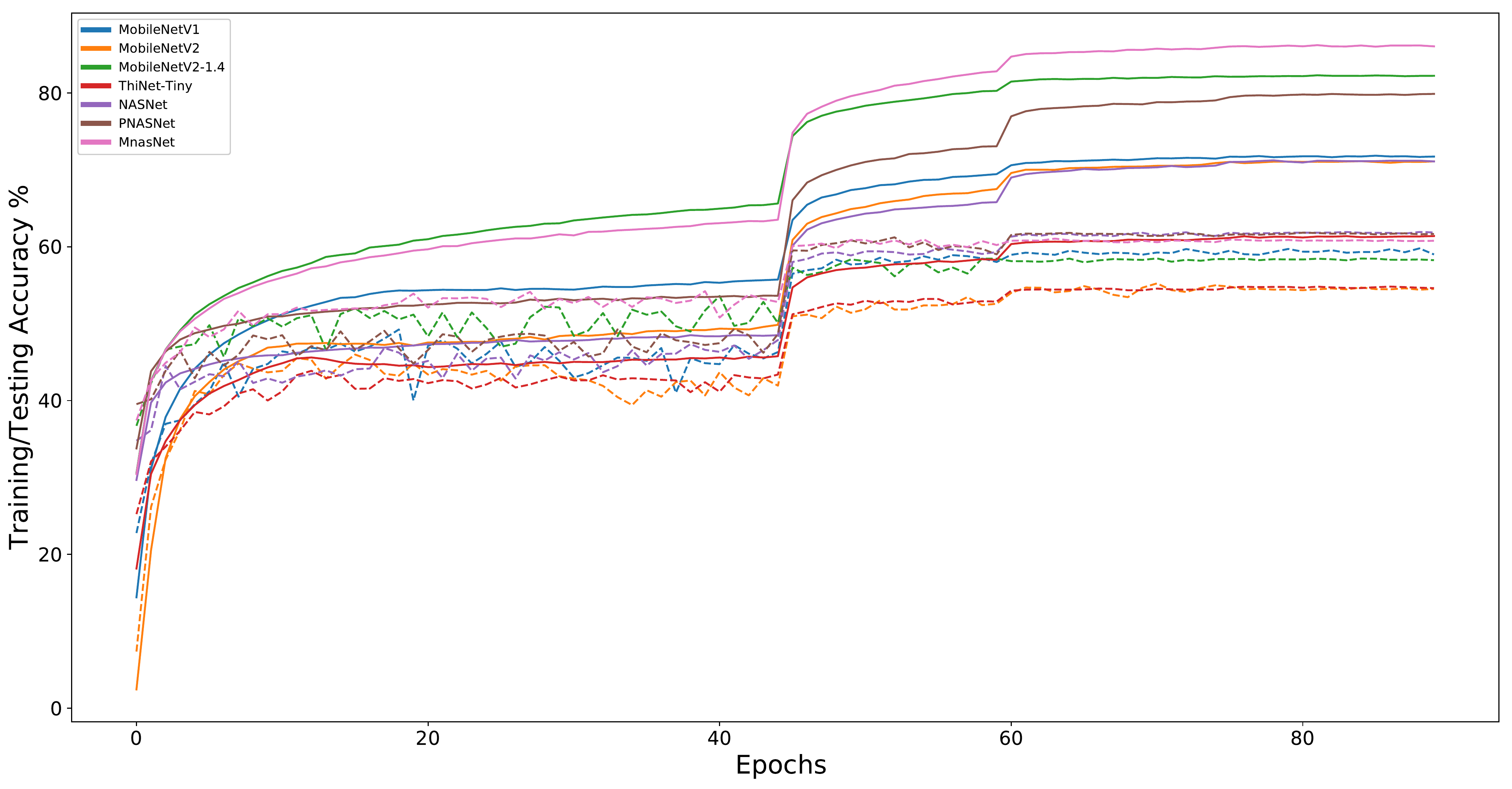}
	\caption{Training and testing accuracy curves on Kinetics for various models of TSN under the problem setting of real-time action recognition. Different color represents different base models. For each model, the solid line represents training accuracy and the dotted line represents testing accuracy. This figure is best viewed in color.}\label{figure: curve}
	
\end{figure*}
\begin{table*}

	\begin{small}
	\begin{center}
		\caption{Number of parameters, FLOPs, mobile CPU latency and accuracy of current compact models on ImageNet and the Kinetics dataset for compact models. The second column represents the number of parameters and the third to seventh column represent FLOPs, accuracy on ImageNet 2012 validation set, single image latency on HUAWEI Mate 10, FLOPs of TSN model and accuracy of the Kinetics validation dataset. For FLOPs calculation, we treat multiply and add operation as a single operation. The last two rows are for reference only.}
		\label{table:final results}
		\setlength{\tabcolsep}{2pt}
		\begin{tabular}{|l|r|r|r|r|r|r|}
			\hline
			Model & Params & FLOPs & ImageNet acc & Image Latency &TSN FLOPs & Kinetics acc\\
			\hline
			ThiNet-Tiny & 1.32M & 1.12B & 57.4\% & 256.03ms &4.64B& 54.6\% \\
			MobileNetV1 & 4.2M & 0.57B & 70.6\% & 106.47ms & 4.56B & 59.0\% \\
			MobileNetV2 & 3.4M & 0.30B & 72.0\% & 89.79ms &2.40B& 54.0\%\\
			MobileNetV2-1.4 & 6.9M & 0.59B & 74.7\% & 144.86ms &4.72B& 58.7\% \\
			NASNet-A-Mobile & 5.3M & 0.57B & 74.0\% & 320.76ms &4.48B& 61.9\% \\
			PNASNet-5-Mobile & 5.1M & 0.59B & 74.2\% & 289.58ms & 4.72B & 61.7\% \\
			MnasNet & 4.2M & 0.32B & 74.2\% & 78.12ms & 2.56B & 60.7\% \\
			\hline 
			BN-Inception & 11.3M &2.04B & 74.8\% & 299.44ms &16.32B& 67.0\%  \\
			ResNet-50 & 25.6M & 3.86B & 75.3\% & 512.70ms &33.88B& 66.8\% \\
			\hline
		\end{tabular}
	\end{center}
\end{small}

\end{table*}

The training curve of all these models are in Fig.~\ref{figure: curve}. According to Table~\ref{table:final results} and Fig.~\ref{figure: curve}, we can easily find that: 
\squishlist
	\item It is interesting that MobileNetV1 has better accuracy than MobileNetV2 on the Kinetics dataset. This observation suggests that some optimized structures on ImageNet like inverse bottleneck may not be suitable for the task of action recognition. This phenomenon shows that designing efficient real-time action recognition models is different from designing image recognition models. 
	\item Although the FLOPs of NASNet-A-Mobile, PNASNet-5-Mobile and MobileNetV2-1.4 are roughly the same, their mobile CPU latency time are totally different. MobileNetV2-1.4 has the least CPU latency while the NASNet-A-Mobile has 2x CPU time on mobile devices. This phenomenon may explained by the fact that MobileNetV2-1.4 only uses simple forward block, while PNASNet-5-Mobile and NASNet-A-Mobile use complex searched branch blocks according to Fig.~\ref{figure: nas}. However, these complex branch blocks improve the accuracy. For example, NASNet-A-mobile achieves the best performance among all these models, and it has the most complex basic building blocks.

	\item MnasNet has the least FLOPs and mobile CPU latency among all these models while having a comparable accuracy. It is a suitable model for further detailed experiments and for deployment on mobile devices.
\squishend
\subsection{Do We Really Need ImageNet?}

Some previous researches such as the SlowFast network~\cite{slowfastarxiv2018,rethinkingarxiv2018} show that in the field of object detection and action recognition, pretrained models on ImageNet is not an essential part. In this section, we will discuss the relationship between ImageNet pretrained models and Kinetics accuracy. Several models are prepared for evaluating the effectiveness of ImageNet.

For models with weights pretrained on ImageNet, we use the same hyperparameter settings as in Sec~4.3. For models trained from scratch, we use longer training epochs to ensure a fair comparison on Kinetics. The learning rate is $0.01$ for first 90 epochs, and reduced to 10\% every 30 epochs. As ~\cite{slowfastarxiv2018} suggests, we use a linear warmup strategy for first 5 epochs.
\begin{table}
	
	\begin{small}
		\begin{center}
			\caption{Results of different pretrained models on the Kinetics dataset. The second column represents whether the model is pretrained on ImageNet and the third column represents accuracy on the Kinetics validation dataset.}
			\label{table:imagenet results}
			\setlength{\tabcolsep}{2pt}
			\begin{tabular}{|l|r|r|}
				\hline
				Model & Pretrained & Kinetics acc\\
				\hline
				ThiNet-Tiny & - & 52.5\% \\
				ThiNet-Tiny & ImageNet & 54.6\% \\  
				\hline	
				MnasNet & - & 57.0\% \\				
				MnasNet & ImageNet & 60.7\% \\		
				\hline	
				PNASNet-5-Mobile & - & 57.0\% \\				
				PNASNet-5-Mobile & ImageNet & 61.7\% \\					
				\hline
				ResNet-50 & - & 63.0\% \\ 
				ResNet-50 & ImageNet & 66.8\% \\	
				\hline
			\end{tabular}
		\end{center}
	\end{small}

\end{table}
According to results in Table~\ref{table:imagenet results}, we can have the following observations In the setting of real-time action recognition, ImageNet pretraining improves 2-3\% accuracy on almost all our base models. This conclusion is different from that in~\cite{slowfastarxiv2018}. This may be explained by the fact that TSN models use global average pooling to simply capture the relationship between frames while SlowFast uses 3D conv to capture the temporal dependencies explicitly. Hence, it is easy for our models to get overfit. ImageNet pretrained models provide a better start point and accelerate the convergence of the training process.

Furthermore, we are interested in the relationship between the accuracy on ImageNet and the accuracy on Kinetics, i.e., does a higher performance on ImageNet directly correlates with a higher Kinetics accuracy?

We perform an ablation study on PNASNet and MnasNet. For pretrained models, we prepare models at different accuracy level, i.e., models at different training epochs on ImageNet for initial weights of PNASNet and MnasNet.
\begin{table}
	
	\begin{small}
		\begin{center}
			\caption{Results of different pretrained models on the Kinetics dataset. The second column represents the accuracy of pretrained models on ImageNet and the third column represents accuracy on the Kinetics validation dataset.}
			\label{table:imagenet results2}
			\setlength{\tabcolsep}{2pt}
			\begin{tabular}{|l|r|r|}
				\hline
				Model & ImageNet acc & Kinetics acc\\
				\hline
				MnasNet & 67.5\%  & 60.3\% \\
				MnasNet  & 74.2\% & 60.7\% \\
				\hline
				PNASNet-5-Mobile  & 68.8\%  & 60.6\% \\
				PNASNet-5-Mobile & 74.2\%  & 61.7\% \\	
				\hline
			\end{tabular}
		\end{center}
	\end{small}
\end{table}
As shown in Table~\ref{table:imagenet results2}, we can see that in the setting of real-time action recognition, ImageNet performance is directly related to Kinetics performance. This phenomenon is also different from previous conclusion~\cite{slowfastarxiv2018}. It may indicate that pretrained models on ImageNet is essential for real-time action recognition, and proves that the linear warmup technique is not sufficient for current models. Moreover, for NAS on real-time action recognition tasks, it is vital to consider the ImageNet factor.
\subsection{Overfitting Matters}

In all our previous models, we set dropout to 0 in our models. Previous researches on ImageNet shows that shallow models do not tend to overfit, i.e., for shallow models, they are easy to underfit. We want to explore this problem in the real-time action recognition setting. Hence, we choose MnasNet as our baseline model and evaluate two kinds of strategies. One is adding dropout before the final classification layer, and the dropout ratio is set to 0.8. The other strategy is partial BN, which fixes all batch normalization layer except the first one. Both strategies are introduced by ~\cite{tsn2016eccv}.
\begin{table}
	
	\begin{small}
		\begin{center}
			\caption{Results of different regularization techniques on the Kinetics dataset. The second column represents pretrained datasets and the third column represents accuracy on the Kinetics validation dataset. Please note that all our models only takes 1 clip from the RGB modality of origin video.}
			\label{table:dropout results}
			\setlength{\tabcolsep}{2pt}
			\begin{tabular}{|l|r|}
				\hline
				Model  & Kinetics acc\\
				\hline
				MnasNet   & 60.7\% \\
				MnasNet + dropout & 61.9\% \\
				MnasNet + fix BN & 62.2\% \\
				\hline
			\end{tabular}
		\end{center}
	\end{small}

\end{table}
As illustrated in Table~\ref{table:dropout results}, we can have the following observation: Different from ImageNet, models on Kinetics are easy to get overfit, even though the MnasNet model is compact. Dropout and fix BN are both effective for migrating the overfitting problem.
\subsection{Extra Modules}

Some researchers propose small modules for improving the performance on both 2D ImageNet and action recognition models. In this section, we choose MnasNet as our baseline model, and use the temporal shift module (TSM)~\cite{tsmarxiv2018} and the squeeze and excitation~(SE) module~\cite{senetcvpr2018} as extra modules. 

The temporal shift module shifts part of the channels along the temporal dimension, which facilitates information exchange among neighboring frames~\cite{tsmarxiv2018} with zero FLOPs cost and negligible time cost. For the SE module, it adaptively recalibrates channel-wise feature responses by explicitly modeling interdependencies between channels~\cite{senetcvpr2018}.

The training hyperparameters are set the same as in Sec~4.1. For TSM, we follow the official guide in~\cite{tsmarxiv2018}. For SE modules, we add them in every block of MnasNet, and we add 10 TSM and SE modules in total. 
\begin{table}
	
	\begin{small}
		\begin{center}
			\caption{Results of different extra models on the Kinetics dataset. The second column represents added extra modalities and the third column represents accuracy on Kinetics validation dataset.}
			\label{table:extra module results2}
			\setlength{\tabcolsep}{2pt}
			\begin{tabular}{|l|r|r|}
				\hline
				Model & Extra module & Kinetics acc\\
				\hline
				MnasNet & /  & 60.7\% \\
				MnasNet & TSM  & 64.6\% \\
				MnasNet & SE  & 62.7\% \\
				MnasNet & TSM + SE + dropout & 66.5\% \\
				\hline
			\end{tabular}
		\end{center}
	\end{small}

\end{table}

As Table~\ref{table:extra module results2} shows, different modules can cooperate together to improve the accuracy on the Kinetics dataset. With the help of these extra modules, MnasNet with TSN can achieve state-of-the-art accuracy with about 6x FLOPs and running times savings, which indicates the effectiveness of our models. 

\subsection{Real-World Applications: A Live Demo}
In this section, we will provide a real-world case study of real-time action recognition on mobile devices. We choose gesture recognition to evaluate our model on real-world scenes. The Jester dataset is a large collection of densely-labeled video clips that show humans performing pre-defined hand gestures in front of a laptop camera or webcam~\cite{20bnjester}. It allows training models to recognize human hand gestures. The dataset has a total number of 148,092 videos and we use the training set with 118,562 for training, and 14,787 in the validation set to evaluate our model.

We use the MnasNet models pretrained from Kinetics and fine-tune it on the target Jester dataset. The fine-tuning is conducted for 25 epochs with initial learning rate 0.001 and decayed by a factor of 0.1 every 10 epochs.

Beyond the simple measurement, we put the model into a live demo on mobile devices with a HUAWEI Mate 10 phone. This application directly captures video frames from cameras and perform action recognition locally, which is an online video understanding scenario. Since TSN process each frame individually, we can cache prediction results of previous frames. Hence, in this section, we will discuss two kinds of inputs: whole sampled clips and sequential input images. Since TSM will shift the feature along the temporal dimension, so it can only be applied in whole video inputs.

\begin{table}
	
	\begin{small}
		\begin{center}
			\caption{Results of different models on the 20BN Jester dataset. The second column represents accuracy on the Jester validation dataset. The third column represents the real CPU latency on HUAWEI MATE 10 when taking the whole 8 sampled frames as input. The fourth column represents the real CPU latency on HUAWEI MATE 10 when the frames are sequentially fed into the action recognition model. For all latencies, we feed the image into action recognition 100 times and report the average result.}
			\label{table:gesture recognition results}
			\setlength{\tabcolsep}{2pt}
			\begin{tabular}{|l|r|r|r|}
				\hline
				Model  & Acc & Batch latency & Sequential latency\\
				\hline 
				MnasNet   & 80.9\% & 533.19ms & 90.24ms\\
				MnasNet+TSM  & 93.6\% & 588.70ms & -\\
				MnasNet+SE & 80.5\% & 558.19ms & 105.09ms\\
				MnasNet+TSM+SE  & 93.7\% & 570.47ms & -\\
				\hline
				ResNet-50~\cite{tsmarxiv2018} & 81.0\% & 3151.57ms & 512.70ms \\
				ResNet-50+TSM~\cite{tsmarxiv2018} & 94.4\% & 3243.11ms & - \\
				\hline
			\end{tabular}
		\end{center}
	\end{small}
\vspace{-0.5cm}
\end{table}

Based on the results in Table~\ref{table:gesture recognition results} and Table~\ref{table:final results}, we can see that our models perform roughly the same as ResNet-50 with only 20\% of its latency and 10\% of its FLOPs. With sequential image inputs, our model can achieve results with about 10FPS on HUAWEI MATE 10, which roughly meets the real-time requirements. To our best knowledge, this is the first paper which applies current deep learning action recognition models into real-world mobile devices and achieves comparable results with state-of-the-art models.
\section{Conclusion}

In this paper, we explored the inference setting of real-time action recognition. Based on this new inference setting, we empirically evaluated current state-of-the-art models. Some suggestions were found through empirical ablation studies including pretrained weights, overfitting issues and extra modules. We achieved state-of-the-art accuracy with 6x FLOPs and inference time saving with the help of extra modules. Finally, a case study on gesture recognition proves that our compact model can be applied in real-time action recognition scenarios on mobile devices. For future work, we will try to design more efficient and effective models in real-time action recognition.
{\small
\bibliographystyle{ieee}
\bibliography{egbib}
}

\end{document}